\documentclass[10pt,twocolumn,letterpaper]{article}

\usepackage[pagenumbers]{cvpr} 

\usepackage[dvipsnames]{xcolor}

\definecolor{cvprblue}{rgb}{0.21,0.49,0.74}
\usepackage[pagebackref,breaklinks,colorlinks,citecolor=cvprblue]{hyperref}

\usepackage{url}
\usepackage{float}
\usepackage{graphicx}
\usepackage{wrapfig}
\usepackage{booktabs}

\def\paperID{9881} 
\def\confName{CVPR}
\def\confYear{2024}

\title{HumanNorm: Learning Normal Diffusion Model for High-quality and Realistic 3D Human Generation}

\author{ Xin Huang\textsuperscript{1\dag,*}, Ruizhi Shao\textsuperscript{2*}, Qi Zhang\textsuperscript{1}, Hongwen Zhang\textsuperscript{2}, Ying Feng\textsuperscript{1}, \\ Yebin Liu\textsuperscript{2}, Qing Wang\textsuperscript{1} \\
$^1$Northwestern Polytechnical University, $^2$Tsinghua University \\
}

\begin{document}
\twocolumn[{%
\renewcommand\twocolumn[1][]{#1}%
\maketitle

\begin{center}
    \centering
    \captionsetup{type=figure}
    \includegraphics[width=\hsize]{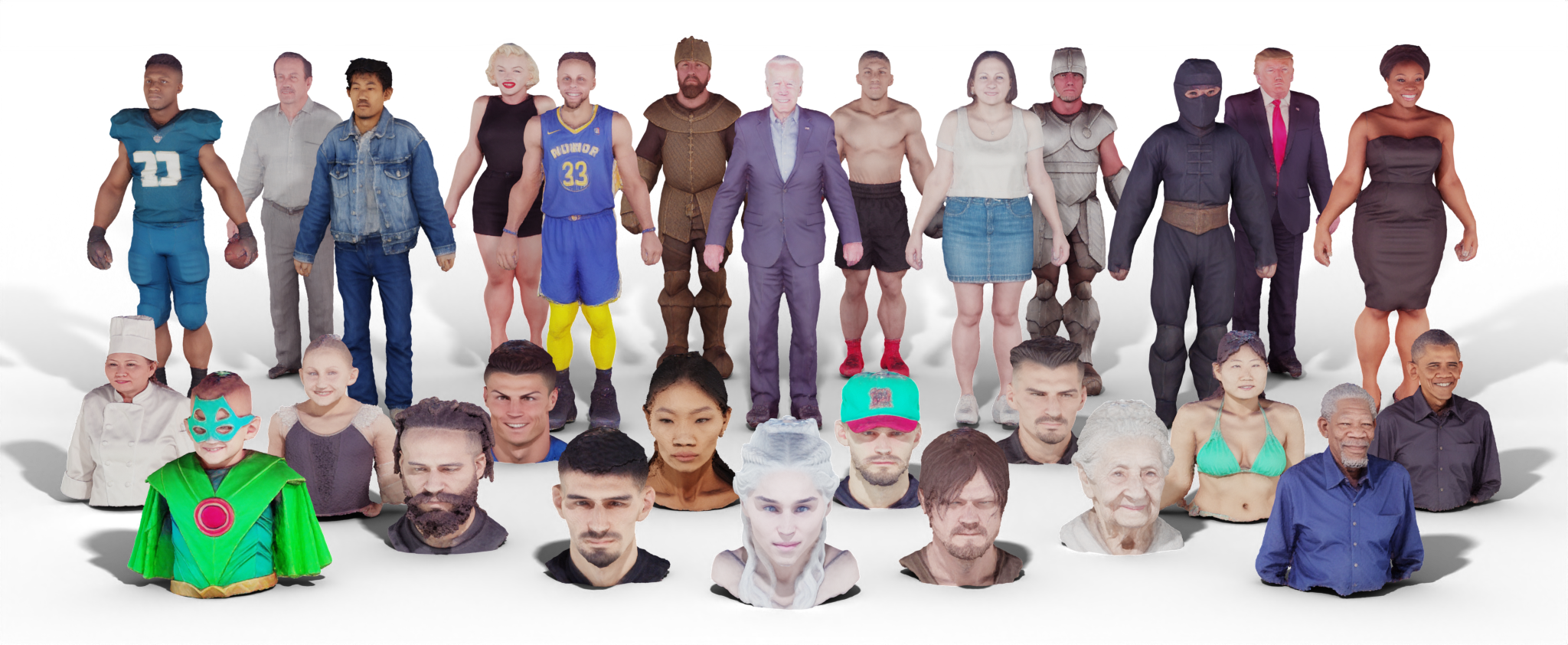}
    \caption{Taking text descriptions as input, HumanNorm has the capability to generate 3D human models with superior geometric quality and realistic textures. The 3D human models produced by HumanNorm can be exported as human meshes and texture maps, making them suitable for  downstream applications.
    }  
    \label{fig:teaser}
\end{center}%
}]

\let\thefootnote\relax\footnotetext{$^\dag$ Work done during an internship at Tsinghua University.}
\let\thefootnote\relax\footnotetext{* Equal contribution.}

\begin{abstract}
Recent text-to-3D methods employing diffusion models have made significant advancements in 3D human generation. However, these approaches face challenges due to the limitations of text-to-image diffusion models, which lack an understanding of 3D structures. Consequently, these methods struggle to achieve high-quality human generation, resulting in smooth geometry and cartoon-like appearances. In this paper, we propose HumanNorm, a novel approach for high-quality and realistic 3D human generation. The main idea is to enhance the model's 2D perception of 3D geometry by learning a normal-adapted diffusion model and a normal-aligned diffusion model. The normal-adapted diffusion model can generate high-fidelity normal maps corresponding to user prompts with view-dependent and body-aware text. The normal-aligned diffusion model learns to generate color images aligned with the normal maps, thereby transforming physical geometry details into realistic appearance. Leveraging the proposed normal diffusion model, we devise a progressive geometry generation strategy and a multi-step Score Distillation Sampling (SDS) loss to enhance the performance of 3D human generation. Comprehensive experiments substantiate HumanNorm’s ability to generate 3D humans with intricate geometry and realistic appearances. HumanNorm outperforms existing text-to-3D methods in both geometry and texture quality. The project page of HumanNorm is \url{https://humannorm.github.io/}.

\end{abstract}    
\section{Introduction}
\label{sec:intro}
Large-scale generative models have achieved significant breakthroughs in diverse domains, including motion~\cite{tevet2023human}, audio~\cite{oord2018parallel,agostinelli2023musiclm}, and 2D image generation~\cite{rombach2022high, nichol2021glide, ramesh2022hierarchical, ramesh2021zero, saharia2022photorealistic}. However, the pursuit of high-quality 3D content generation~\cite{poole2023dreamfusion,chen2023fantasia3d,shi2023mvdream,sun2023dreamcraft3d} following the success of 2D generation poses a novel and meaningful challenge. Within the broader scope of 3D content creation, 3D human generation~\cite{hong2022avatarclip,kolotouros2023dreamhuman,liao2023tada} holds particular significance. It plays a pivotal role in applications such as AR/VR, holographic communication, and the metaverse. 

To achieve 3D content generation, a straightforward approach is to train generative models like GANs or diffusion models to generate 3D representations~\cite{chan2022efficient,an2023panohead,wang2023rodin,hu2023humanliff}. However, these approaches face challenges due to the scarcity of current 3D datasets, resulting in restricted diversity and suboptimal generalization. 
To overcome these challenges, recent methods~\cite{poole2023dreamfusion,lin2023magic3d,metzer2023latent} adopt a 2D-guided approach to achieve 3D generation. Their core framework builds upon pre-trained text-to-image diffusion models and distills 3D contents from 2D generated images through Score Distillation Sampling (SDS) loss~\cite{poole2023dreamfusion}. Leveraging the image generation priors learned from large-scale datasets, this framework enables more diverse 3D generation. However, current text-to-image diffusion models primarily emphasize the generation of natural RGB images, which results in a limited perception of 3D geometry structure and view direction. This limitation can result in Janus (multi-faced) artifacts and smooth geometry. Moreover, the texture of the 3D contents generated by existing methods is sometimes not based on geometry, which can result in fake 3D details, particularly in wrinkles and hair. Although some 3D human generation methods~\cite{cao2023dreamavatar,kolotouros2023dreamhuman,liao2023tada} introduce human body models such as SMPL~\cite{loper2023smpl} for animation and enhancing the quality of body details, they fail to address these fundamental limitations. Their results still suffer from sub-optimal geometry, fake 3D details and over-saturated texture.

\begin{figure*}[t]
\includegraphics[width=\textwidth]{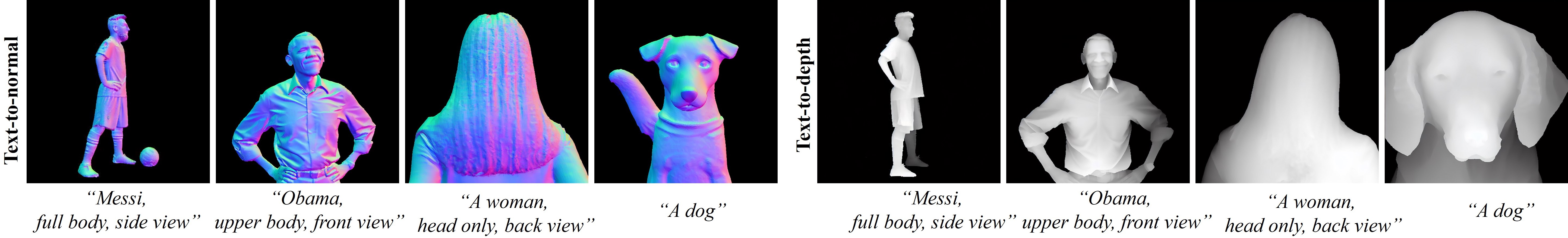}
\vspace{-15pt}
\caption{\textbf{2D results by normal-adapted and depth-adapted diffusion models.} The view-dependent texts like ``front view" are utilized to control the view direction. The body-aware texts like ``upper body" are employed to control which body part is generated. }
\vspace{-10pt}
\label{fig:2Dgeneration}
\end{figure*}

In this paper, we present HumanNorm, a novel approach for generating high-quality and realistic 3D human models. The core idea is introducing a normal diffusion model to enhance the perception of 2D diffusion model for 3D geometry. HumanNorm is divided into two components: geometry generation and texture generation. For the geometry generation, we train a \textit{normal-adapted diffusion model} using multi-view normal maps rendered from 3D human scans and prompts with view-dependent and body-aware text. Compared with text-to-image diffusion models, the normal-adapted diffusion model filters out the influence of texture and can generate high-fidelity surface normal maps according to prompts. This ensures the generation of 3D geometric details and avoids Janus artifacts. Since normal maps lack depth information, we also learn a depth-adapted diffusion model to further enhance the perception of 3D geometry. The 2D results generated by these diffusion models are presented in \cref{fig:2Dgeneration}. The geometry is generated using both normal and depth SDS losses, which are based on our normal-adapted and depth-adapted diffusion models. Furthermore, a progressive strategy is designed to reduce geometric noise and enhance geometry quality.

\begin{figure}[!t]
\includegraphics[width=\linewidth]{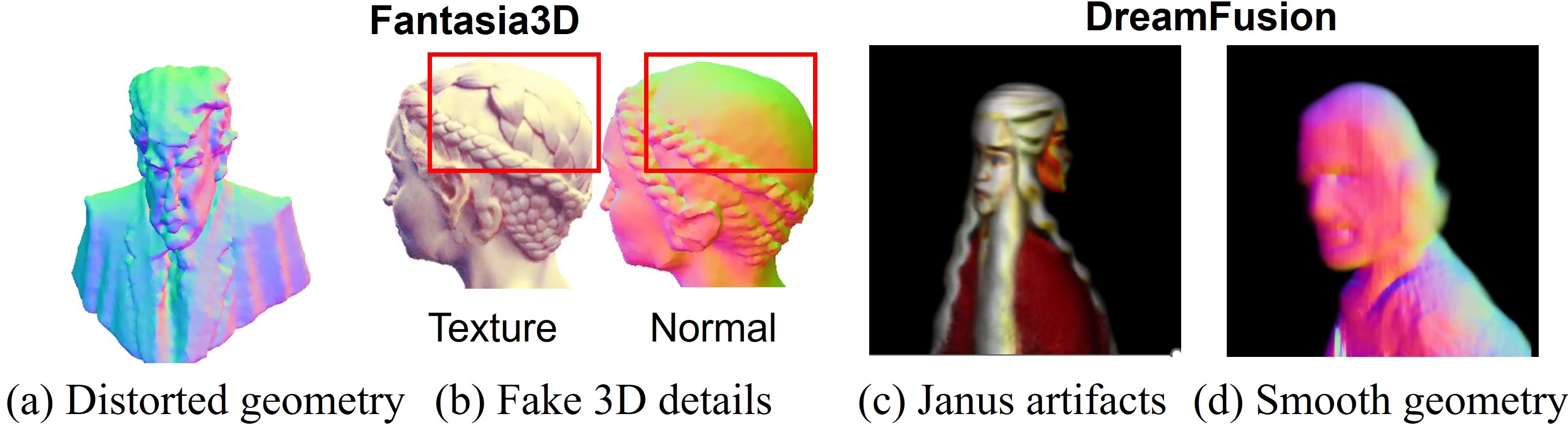}
\vspace{-15pt}
\caption{\textbf{Problems of existing methods.} }
\label{fig:artifacts}
\vspace{-10pt}
\end{figure}

As previously discussed, the core challenges for texture generation are fake 3D details and over-saturated appearances, as illustrated in \cref{fig:artifacts}. To avoid fake 3D details, we learn a \textit{normal-aligned diffusion model} from normal-image pairs. This model efficiently integrates human geometric information into the texture generation process by taking normal maps as conditions. It accounts for elements such as shading caused by geometric folds and aligns the generated texture with surface normal. To tackle the over-saturated appearances, we introduce a multi-step SDS loss based on our normal-aligned diffusion model for texture generation. The loss recovers images with multiple diffusion steps, ensuring a more natural appearance of the generated texture.

The 3D models generated by HumanNorm are presented in ~\cref{fig:teaser}. The key contributions of this paper are: 
\begin{enumerate}
    \setlength\itemsep{0pt}
    \item We propose a method for detailed human geometry generation by introducing a normal-adapted diffusion model that can generate normal maps from prompts with view-dependent and body-aware text.
    \item We propose a method for geometry-based texture generation by learning a normal-aligned diffusion model, which transforms physical geometry details into realistic appearances.  
    \item We introduce the multi-step SDS loss to mitigate over-saturated texture and a progressive strategy for enhancing stability in geometry generation.
\end{enumerate}

\section{Related work}
Our study is primarily centered on the realm of text-to-3D, with a specific emphasis on text-to-3D human generation. Here, we revisit some recent work related to our method.

\textbf{Text-to-3D content generation.} Early methods, such as CLIP-Forge~\cite{sanghi2022clip}, DreamFields~\cite{jain2022zero}, and CLIP-Mesh~\cite{mohammad2022clip}, combine a pre-trained CLIP~\cite{radford2021learning} model with 3D representations, and generate 3D content under the supervision of CLIP loss. DreamFusion~\cite{poole2023dreamfusion} introduces the SDS loss and generates NeRF~\cite{mildenhall2021nerf} under the supervision of a text-to-image diffusion model. Following this, Magic3D~\cite{lin2023magic3d} proposes a two-stage method that employs both NeRF and mesh for high-resolution 3D content generation. Latent-NeRF~\cite{metzer2023latent} optimizes NeRF in the latent space using a latent diffusion model to avoided the burden of encoding images. TEXTure~\cite{richardson2023texture} introduces a method for texture generation, transfer, and editing. Fantasia3D~\cite{chen2023fantasia3d} decomposes the generation process into geometry and texture generation to enhance the performance of 3D generation. To address the over-saturation issue, ProlificDreamer~\cite{wang2023prolificdreamer} proposes a Variational Score Distillation (VSD) loss to produce high-quality NeRF. IT3D~\cite{chen2023it3d} introduces GAN loss and leverages generated 2D images to enhance the quality of 3D contents. MVDream~\cite{shi2023mvdream} proposes a multi-view diffusion model to generate consistent multi-views for 3D generation. DreamGaussian~\cite{tang2023dreamgaussian} uses 3D Gaussian splatting~\cite{kerbl20233d} to accelerate the generation process. However, these methods are unable to generate high-quality 3D humans, leading to Janus artifacts and unreasonable body proportions. Our method addresses these issues by introducing normal-adapted diffusion model that can generate normal maps from prompts with view-dependent and body-aware text.



\textbf{Text-to-3D human generation.} Recently, EVA3D~\cite{hong2023eva3d}, LSV-GAN~\cite{xu2023efficient}, GETAvatar~\cite{zhang2023getavatar}, Get3DHuman~\cite{xiong2023get3dhuman} introduce GAN-based frameworks to directly generate 3D representations for 3D human generation. AvatarCLIP~\cite{hong2022avatarclip} integrates SMPL and Neus~\cite{wang2021neus} to create 3D humans, leveraging CLIP for a supervision. DreamAvatar~\cite{cao2023dreamavatar} and AvatarCraft~\cite{jiang2023avatarcraft} utilize the pose and shape of the parametric SMPL model as a prior, guiding the generation of humans. DreamWaltz~\cite{huang2023dreamwaltz} creates 3D humans using a parametric human body prior, incorporating 3D-consistent occlusion-aware SDS and 3D-aware skeleton conditioning. DreamHuman~\cite{kolotouros2023dreamhuman} generates animatable 3D humans by introducing a pose-conditioned NeRF that is learned using imGHUM. AvatarBooth~\cite{zeng2023avatarbooth} uses dual fine-tuned diffusion models separately for the human face and body, enabling the creation of personalized humans from casually captured face or body images. The most recent model, AvatarVerse~\cite{zhang2023avatarverse}, trains a ControlNet with DensePose~\cite{guler2018densepose} as conditions to enhance the view consistency of 3D human generation. TADA~\cite{liao2023tada} derives SMPL-X~\cite{pavlakos2019expressive} with a displacement layer and a texture map, using hierarchical rendering with SDS loss to produce 3D humans. While these methods reduce Janus artifacts and unreasonable body shapes by introducing human body models, they still produce 3D humans with fake 3D details, over-saturation and smooth geometry. Moreover, the introduction of SMPL presents challenges for these methods in generating 3D humans with intricate clothing such as puffy skirts and hats. Our method addresses these issues by learning normal diffusion model and introducing multi-step SDS loss, thereby enhancing the both geometry and texture quality of 3D humans.




\begin{figure*}[!t]
\includegraphics[width=\textwidth]{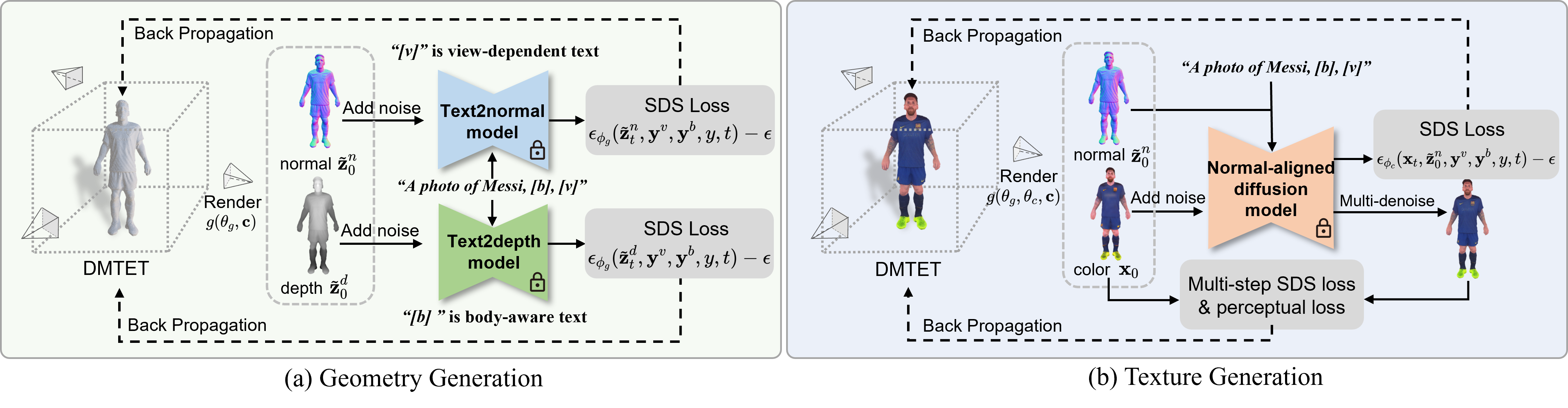}
\vspace{-20pt}
\caption{\textbf{Overview of HumanNorm.} Our method is designed for high-quality and realistic 3D human generation from given prompts. The whole framework consists of geometry and texture generation. We first propose the normal-adapted and depth-adapted diffusion model for the geometry generation. These two models can guide the rendered normal and depth maps to approach the learned distribution of high-fidelity normal and depth maps through the SDS loss, thereby achieving high-quality geometry generation. In terms of texture generation, we introduce the normal-aligned diffusion model. The normal-aligned diffusion model leverages normal maps as guiding cues to ensure the alignment of the generated texture with geometry. We first exclusively employ the SDS loss and then incorporate the multi-step SDS and perceptual loss to achieve realistic texture generation.}
\vspace{-5pt}
\label{fig:pipeline}
\end{figure*}

\section{Preliminary}
\label{sec:preliminary}

\subsection{Diffusion-guided 3D Generation Framework}
\label{sec:current_diffusion_framework}
When provided with text $y$ as the generation target, the core of the diffusion-guided 3D generation framework aims to align the images $\mathbf{x}_0$ rendered from the 3D representation $\theta$ with the generated image distribution $p(\mathbf{x}_0|y)$ of the 2D diffusion model. Specifically, during the 3D generation process, the rendered images  $\mathbf{x}_0$ are obtained by randomly sampling cameras $\mathbf{c}$ and rendering through a differentiable rendering function $g(\theta, \mathbf{c})$. Suppose the rendered images from various angles are distributed as $q^{\theta}(\mathbf{x}_0|y)=\int q^{\theta}(\mathbf{x}_0|y, \mathbf{c})p(\mathbf{c})d\mathbf{c}$, the optimization objective of diffusion-guided 3D generation framework can be represented as follows:
\begin{equation}
\label{equ:direct}
    \min_{\theta} D_{KL}(q^{\theta}(\mathbf{x}_0|y) \ \| \  p(\mathbf{x}_0|y)).
\end{equation}
Directly optimizing this objective is highly challenging, and recent methods have proposed losses such as SDS~\cite{poole2023dreamfusion} and VSD~\cite{wang2023prolificdreamer} to solve it. To further enhance the quality of geometry, Fantasia3D~\cite{chen2023fantasia3d} proposes to disentangle the geometry $\theta_g$ and appearance $\theta_c$ in the 3D representation $\theta$. In the geometry stage, it aligns $q^{\theta_g}(\mathbf{z}_0^n|y)$, the distribution  of the rendered normal maps $\mathbf{z}_0^n$, with the natural image distribution $p(\mathbf{x}_0|y)$:
\begin{equation}
    \min_{\theta_g} D_{KL}(q^{\theta_g}(\mathbf{z}_0^n|y) \ \| \ p(\mathbf{x}_0|y)).
\end{equation}
In the texture stage, the texture of 3D objects is optimized through \cref{equ:direct}.

\subsection{Bottleneck of Diffusion-guided 3D Generation}
The bottleneck of the diffusion-guided 3D generation lies in the T2I (text-to-image) diffusion model, which confines itself to parameterize the probability distribution of natural RGB images, denoted as $p(\mathbf{x}_0|y)$. Therefore, current T2I diffusion model lacks the understanding of both view direction and geometry. Consequently, 3D generation directly guided by the T2I diffusion model (\cref{equ:direct}) leads to Janus artifacts and low-quality geometry as shown in \cref{fig:artifacts}~(c-d). Although Fantasia3D disentangles geometry and texture, it still encounters issues originating from the T2I diffusion model in both geometry and texture stages. In the geometry stage, directly aligning the rendered normal maps distribution $q^{\theta_g}(\mathbf{z}_0^n|y)$ with the natural images distribution $p(\mathbf{x}_0|y)$  is inappropriate since normal maps significantly differ from RGB images. This alignment results in geometry distortions and artifacts, as depicted in ~\cref{fig:artifacts} (a). In the texture stage, minimizing the divergence between the appearance distribution $q^{\theta_c}(\mathbf{x}_0|y)$ and the natural image distribution $p(\mathbf{x}_0|y)$ may lead to fake 3D details due to the absence of geometric guidance, as presented in ~\cref{fig:artifacts}~(b).

\section{Method}
We propose HumanNorm to achieve high-quality and realistic 3D human generation. The whole generation framework has a geometry stage and a texture stage, as shown in ~\cref{fig:pipeline}. In this section, we first introduce our normal diffusion model, which consists of a normal-adapted diffusion model and a normal-aligned diffusion model (~\cref{sec:geo_aware}). Then in the geometry stage, based on the normal-adapted diffusion model, we utilize the DMTET~\cite{shen2021deep} as the 3D representation and propose a progressive generation strategy to achieve high-quality geometry generation (~\cref{sec:geo_stage}). In texture stage, building upon the normal-aligned diffusion model, we propose the multi-step SDS loss for high-fidelity and realistic appearance generation (~\cref{sec:tex_stage}).

\subsection{Normal Diffusion Model}
\label{sec:geo_aware}
In the pursuit of generating a high-quality and realistic 3D human from a given text target $y$, the first challenge lies in achieving precise geometry generation. This entails aligning the distributions of rendered normal maps $q^{\theta_g}(\mathbf{z}_0^n|\mathbf{c}, y)$ from multiple viewpoints $\mathbf{c}$ with an ideal normal maps distribution $\hat{p}(\mathbf{z}_0^n|\mathbf{c}, y)$. The next challenge is to generate the realistic texture $\theta_c$ while ensuring its coherence with the established geometry $\theta_g$. Therefore, minimizing the divergence between the distribution of rendered images $q^{\theta_c}(\mathbf{x}_0|\mathbf{c}, y)$ and an ideal geometry-aligned images distribution $\hat{p}(\mathbf{x}_0|\mathbf{c}, \theta_g, y)$ becomes essential. The ideal optimization objective is formulated as follows:
\begin{equation}
\begin{gathered}
    \min_{\theta_g, \theta_c} \underbrace{ D_{KL}(q^{\theta_g}(\mathbf{z}_0^n| \mathbf{c}, y) \ \| \ \hat{p}(\mathbf{z}_0^n|\mathbf{c}, y))}_{geometry\ generation\ objective} \\
    + \underbrace{D_{KL}(q^{\theta_c}(\mathbf{x}_0|\mathbf{c}, y) \ \| \ \hat{p}(\mathbf{x}_0|\mathbf{c}, \theta_g, y))}_{texture\ generation\ objective}.
\end{gathered}
\end{equation}

However, as discussed in \cref{sec:current_diffusion_framework}, the existing T2I (text-to-image) diffusion model is limited to parameterize the distribution of natural RGB images, denoted as $p(\mathbf{x}_0|y)$, which deviates significantly from the ideal distributions $\hat{p}(\mathbf{z}_0^n|\mathbf{c}, y)$ and $\hat{p}(\mathbf{x}_0|\mathbf{c}, \theta_g, y)$. To bridge this gap, we propose the incorporation of normal maps, representing the 2D perception of human geometry, into the T2I diffusion model to approximate $\hat{p}(\mathbf{z}_0^n|\mathbf{c}, y)$ and $\hat{p}(\mathbf{x}_0|\mathbf{c}, \theta_g, y)$. For the geometry component, we propose to fine-tune the diffusion model, adapting it to generate the distribution of normal map $p(\mathbf{z}_0^n|y)$. In the context of texturing, we utilize normal maps $\mathbf{z}_0^n$ as conditions to guide the diffusion model $p(\mathbf{x}_0|\mathbf{z}_0^n, y)$ in generating normal-aligned images, which ensures that the generated texture aligns with the geometry. 
In addition, we further introduce view-dependent text $\mathbf{y}^v$ (\textit{e.g. ``front view"}) and body-aware text $\mathbf{y}^b$ (\textit{e.g. ``upper body"}), serving as an additional condition for the diffusion model. This strategy ensures that the generated images align with the view direction and enables body part generation, as depicted in \cref{fig:2Dgeneration}. The final optimization objective is:
\begin{equation}
\label{equ:orientation}
\begin{gathered}
    \min_{\theta_g, \theta_c} D_{KL}(q^{\theta_g}(\mathbf{z}_0^n| \mathbf{c}, y) \ \| \ p(\mathbf{z}_0^n|\mathbf{y}^v, \mathbf{y}^b, y)) + \\
    D_{KL}(q^{\theta_c}(\mathbf{x}_0|\mathbf{c}, y) \ \| \ p(\mathbf{x}_0|\mathbf{z}_0^n, \mathbf{y}^v, \mathbf{y}^b, y)).
\end{gathered}
\end{equation}
Next, we will introduce our 3D human generation framework and construction of the normal-adapted diffusion model and normal-aligned diffusion model used to parameterize $p(\mathbf{z}_0^n|\mathbf{y}^v, \mathbf{y}^b, y)$ and $p(\mathbf{x}_0|\mathbf{z}_0^n, \mathbf{y}^v, \mathbf{y}^b, y)$ for geometry and texture generation.

\begin{figure*}[!th]
\includegraphics[width=\textwidth]{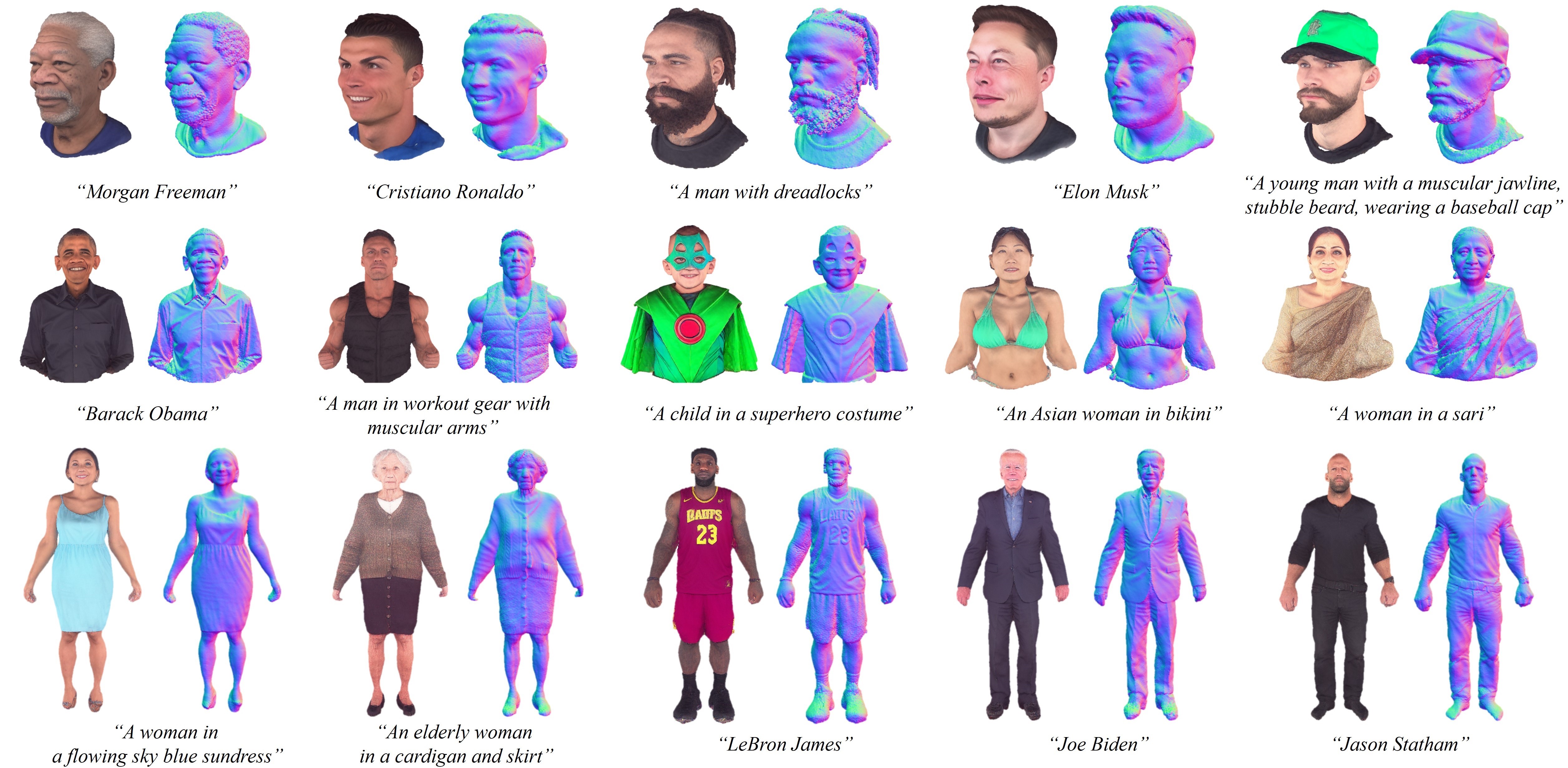}
\caption{\textbf{Examples of 3D humans generated by HumanNorm.} A single view and the corresponding normal map are rendered for visualization. \textbf{See supplementary for video results}.}
\label{fig:main_results}
\end{figure*}

\subsection{Geometry Generation} \label{sec:geo_stage}
\subsubsection{Normal-adapted Diffusion Model} 
Constructing the normal-adapted diffusion model for high-quality geometry generation faces several challenges. First, existing 3D human datasets are scarce, leading to a limited number of normal maps for training. Therefore, we employ a fine-tuning strategy to adapt a text-to-image diffusion model into a text-to-normal diffusion model. Then we find the rendered normal maps undergo dramatic changes with variations in viewing angles, which results in potential overfitting or underfitting issues. To mitigate this effect and encourage the diffusion model to focus on perceiving the details of geometry, we transform the normal maps $\mathbf{z}_0^n$ from the world coordinate to camera coordinates by the rotation $R$ of the camera parameters. The transformed normal maps $\tilde{\mathbf{z}}_0^n$ are used for training of the normal-adapted diffusion model.
As discussed in \cref{sec:geo_aware}, we add the view-dependent text $\mathbf{y}^v$ and body-aware text $\mathbf{y}^b$ as addition conditions. The fine-tuning process employs this optimization objective:
\begin{equation}
    \min_{\phi_g} \mathbb{E}_{\mathbf{c}, t, \epsilon} 
    \left[ \|\mathbf{\epsilon}_{\phi_g}(\alpha_t\tilde{\mathbf{z}}_0^n+\sigma_t, \mathbf{y}^v, \mathbf{y}^b, y, t) - \epsilon\|_2^2 \right],
\end{equation}
where $\mathbf{c}$ is a camera pose, $t$ is a timestep, $\epsilon$ denotes noise and $y$ is a prompt.   $\sigma_t$ and $\alpha_t$ are the parameters of the diffusion scheduler. $\mathbf{\epsilon}_{\phi_g}(\cdot)$ is the normal-adapted diffusion model.

SDS loss~\cite{poole2023dreamfusion} is widely employed in various diffusion-guided 3D generation frameworks. It translates the optimization objective in \cref{equ:direct} into the optimization of the divergence between two distributions with diffusion noise, thereby achieving 3D generation. Our geometry is optimized by the normal SDS loss based on the trained normal-adapted diffusion model:
\begin{equation} 
\begin{gathered}
    \nabla \mathcal{L}_{SDS}(\theta_g) = \\
    \mathbb{E}_{\mathbf{c}, t, \epsilon} \left[ \omega(t)(\mathbf{\epsilon}_{\phi_g}(\tilde{\mathbf{z}}_t^n, \mathbf{y}^v, \mathbf{y}^b, y, t) - \epsilon)\frac{\partial g(\theta_g, \mathbf{c})}{\partial \theta_g}  \right].
\end{gathered}
\end{equation}
where $\tilde{\mathbf{z}}_t^n$ corresponds to the rendered normal map $\tilde{\mathbf{z}}_t^0$ with the noise $\epsilon$ at timestep $t$. $\omega(t)$ is the parameters of the diffusion scheduler. $g(\theta_g, \mathbf{c})$ denotes render the normal map at camera pose $\mathbf{c}$ from geometry $\theta_g$. In addition to normal SDS loss, we also fine-tune a depth-adapted diffusion model by simply changing normal maps to depth maps to calculate depth SDS loss. We found the depth SDS loss can reduce geometry distortion and artifacts in geometry generation, as shown in \cref{fig:ablation_text2depth}.


\subsubsection{Progressive Geometry Generation}
\label{sec:progressive_geometry} 
DMTET~\cite{shen2021deep} is used as our 3D representation. To augment the robustness of 3D human generation, we initialize it with a neutral body mesh. We propose a progressive strategy including progressive positional encoding and progressive SDF loss to mitigate geometric noise and enhance the overall quality of geometry generation. 

Positional encoding~\cite{mildenhall2021nerf,muller2022instant} maps each component of input vectors to a higher-dimensional space, thereby enhancing the 3D representation's ability to capture high-frequency details. However, we found that the high frequency of positional encoding can also lead to noisy surface. This is due to the DMTET prioritizing coarse geometry during the initial optimization stage, resulting in the failure to translate high-frequency input into geometric details. To solve this, we employ a mask to suppress high-frequency components of positional encoding for SDF function in DMTET during the initial stage. This allows the network to focus on low-frequency components of geometry and improving the training stability in the beginning. As training progresses, we gradually reduce the mask for high-frequency components. Thereby enhancing the details such as clothes wrinkle.

In addition, the progressive SDF loss is introduced to further improve the quality of geometry generation. We first record the SDF functions of DMTET before reducing the high-frequency mask, denoted as $\mathbf{s}(x)$. Then as training progresses, we add the SDF loss to mitigate strange geometry deformations: 
\begin{equation}
    \mathcal{L}_{SDF}(\theta_g) = \sum_{x \in P}\|\tilde{\mathbf{s}}_{\theta_g}(x) - \mathbf{s}(x) \|_2^2,
\end{equation}
where $\tilde{\mathbf{s}}_{\theta_g}(x)$ is the SDF function in DMTET and $P$ is the set of random sampling points. This strategy can effectively avoid unreasonable body proportions.

\subsection{Texture Generation}
\label{sec:tex_stage}
\subsubsection{Normal-aligned Diffusion Model}
In texture generation, we fix the geometry parameters $\theta_g$ and introduce the normal-aligned diffusion model as guidance. The normal-aligned diffusion model can translate physical geometry details into realistic appearance and ensure the generated texture is aligned with the geometry. Specifically, we employ the strategy of ControlNet~\cite{zhang2023adding} to incorporate transformed normal maps $\tilde{\mathbf{z}}_0^n$ as the guided condition of the T2I diffusion model. The training objective of the normal-aligned diffusion model is as follows:
\begin{equation}
   \min_{\phi_c} \mathbb{E}_{\mathbf{c}, t, \epsilon} 
   \left[ \|\mathbf{\epsilon}_{\phi_c}(\alpha_t\mathbf{x}_0+\sigma_t, \tilde{\mathbf{z}}_0^n, \mathbf{y}^v, \mathbf{y}^b, y, t) - \epsilon\|_2^2 \right] 
\end{equation}
After training, we propose a multi-step SDS loss based on the normal-aligned diffusion model for photo-realistic texture generation.

\subsubsection{Multi-step SDS Loss}
\label{sec:coarse_to_fine}
We generate texture in two stages. In the initial stage, we employ the vanilla SDS loss of the normal-aligned diffusion model $\mathbf{\epsilon}_{\phi_c}$ for texture generation:
\begin{equation}
\begin{gathered}
    \nabla \mathcal{L}_{SDS}(\theta_c) = \\
    \mathbb{E}_{\mathbf{c}, t, \epsilon} \left[ \omega(t)(\mathbf{\epsilon}_{\phi_c}(\mathbf{x}_t, \tilde{\mathbf{z}}_0^n, \mathbf{y}^v, \mathbf{y}^b, y, t) - \epsilon)\frac{\partial g(\theta_c, \mathbf{c})}{\partial \theta_c}  \right].
\end{gathered}
\end{equation}
While SDS loss can lead to over-saturated styles and appear less natural as shown in ~\cref{fig:ablations1}~(c), it efficiently optimizes a reasonable texture as an initial value. We subsequently refine the texture through multi-step SDS and perceptual loss. Different from SDS loss, multi-step SDS loss needs multiple diffusion steps to recover the distribution of RGB images, which promotes stability during optimization and avoids getting trapped in local optima. As a result, the generated images appear more natural. To further prevent over-saturation effects, the perceptual loss is also applied to keep the natural style of the rendering images consistent with the images generated by the normal-aligned diffusion model. The loss is defined as:
\begin{equation}
\small
\begin{gathered}
    \nabla \mathcal{L}_{MSDS}(\theta_c) \approx \\
    \mathbb{E}_{ \mathbf{c}, t, \epsilon}\left[ \omega(t)(h(\mathbf{x}_t, \tilde{\mathbf{z}}_0^n, \mathbf{y}^v, \mathbf{y}^b, y, t) - \mathbf{x}_0) \frac{\partial g(\theta_c, \mathbf{c})}{\partial \theta} \right] +  
    \lambda_p \mathbb{E}_{\mathbf{c}, t, \epsilon} \\
    \left[  \left(V(h(\mathbf{x}_t, \tilde{\mathbf{z}}_0^n, \mathbf{y}^v, \mathbf{y}^b, y, t)) - V(\mathbf{x}_0)\right)\frac{\partial V(\mathbf{x}_0)}{\partial \mathbf{x}_0}\frac{\partial g(\theta_c, \mathbf{c})}{\partial \theta_c} \right], 
\end{gathered}
\end{equation}
where $V$ is the first $k$ layers of the VGG network~\cite{simonyan2015very}. $h(\mathbf{x}_t, \tilde{\mathbf{z}}_0^n, \mathbf{y}^v, \mathbf{y}^b, y, t)$ denotes the multi-step image generation function of the normal-aligned diffusion model. $\lambda_p$ is the weights of perceptual loss.

\section{Experiment}



\begin{figure*}[t]
\includegraphics[width=\textwidth]{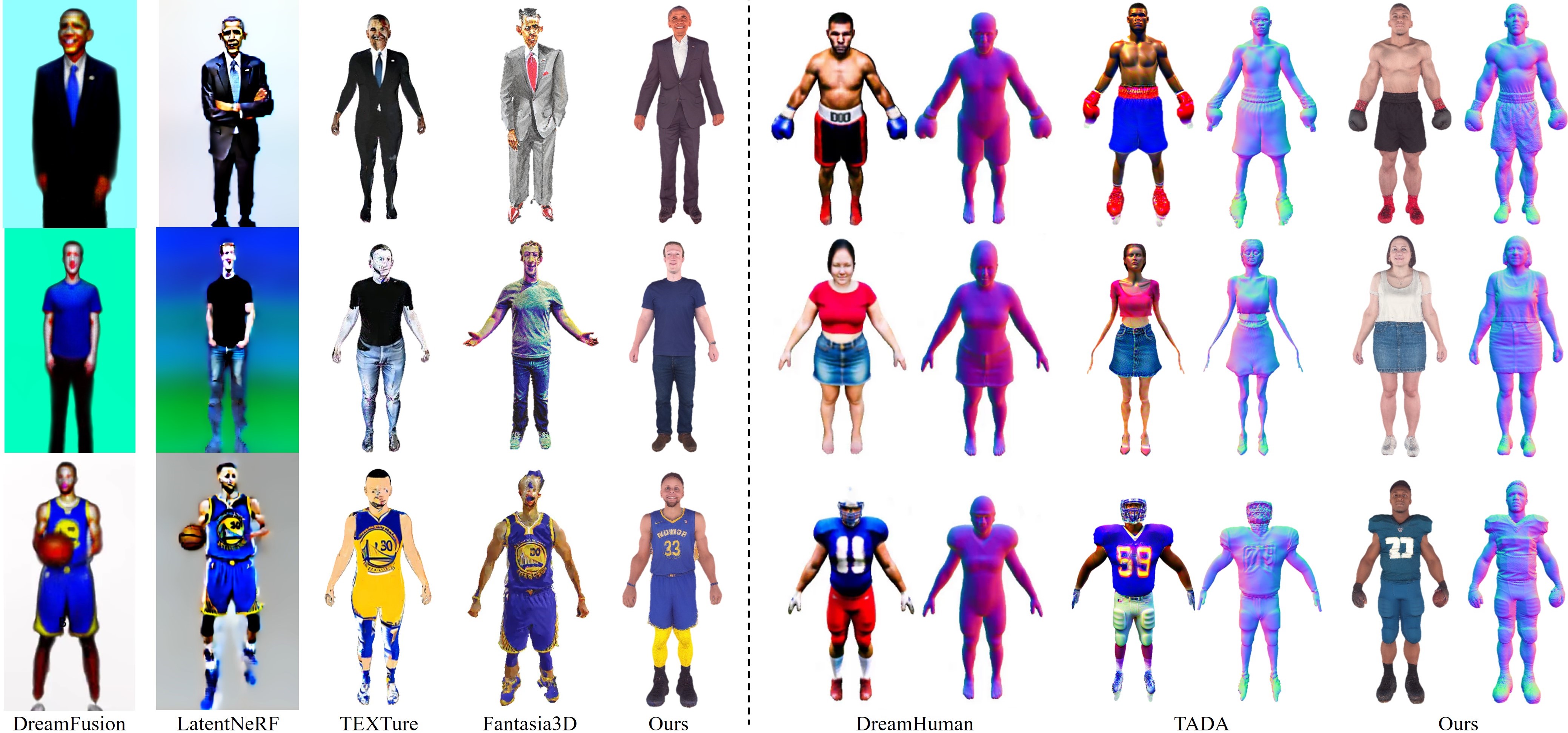}
\vspace{-15pt}
\caption{\textbf{Comparisons with text-to-3D content methods and text-to-3D human methods.} The results of DreamFusion are generated by unofficial code. The results of DreamHuman are taken from its original paper and project page.}
\vspace{-10pt}
\label{fig:comp}
\end{figure*}

\subsection{Implementation Details}
For each prompt, our method needs 15K iterations for geometry generation and 10K iterations for texture generation. The entire generation process takes about 2 hours on a single NVIDIA RTX 3090 GPU with 24 GB memory. The final rendered images and videos have a resolution of $1024\times1024$. Additional details, including dataset, training settings, and more, can be found in our supplementary.

\subsection{Qualitative Evaluation}

The examples of 3D humans generated by HumanNorm is shown in \cref{fig:main_results}. Furthermore, we present qualitative comparisons with text-to-3D content methods including DreamFusion~\citep{poole2023dreamfusion}, LatentNeRF~\citep{metzer2023latent}, TEXTure~\citep{richardson2023texture}, and Fantasia3D~\citep{chen2023fantasia3d}, as well as text-to-3D human methods including DreamHuman~\citep{kolotouros2023dreamhuman} and TADA~\citep{liao2023tada}. 


\noindent\textbf{Comparison with text-to-3D content methods.} As illustrated in ~\cref{fig:comp}, the results produced by text-to-3D content methods present some challenges. The proportions of the generated 3D humans tend to be distorted, and the texture appears to be over-saturated and noisy. DreamFusion struggles to generate full-body humans, often missing the feet, even given a prompt like ``the full body of...''. In contrast, our method delivers superior results with more accurate geometry and realistic textures.

\noindent\textbf{Comparison with text-to-3D human methods.} As shown in ~\cref{fig:comp}, text-to-3D human methods yield outcomes with enhanced geometry due to the integration of SMPL-X and imGHUM human body models. In contrast, HumanNorm can create 3D humans with a higher level of geometric detail, such as wrinkles in clothing and distinct facial features. Furthermore, text-to-3D human methods also encounter issues with over-saturation, while our method can generate more lifelike appearances thanks to the multi-step SDS loss.

\begin{table}
  \centering
  \begin{tabular}{@{}lcc@{}}
    \toprule
    Method & FID $\downarrow$ &CLIP Score $\uparrow$\\
    \midrule
    DreamFusion & 145.2 & 28.65\\
    LatentNeRF & 152.6 & 27.42\\
    TEXTure & 142.8 & 27.08\\
    Fantasia3D & 120.6 & 28.47\\
    \midrule
    DreamHuman & 111.3 &30.15 \\
    TADA & 120.0  &30.65 \\
    \midrule
    HumanNorm (Ours) & 92.5 & 31.70\\
    \bottomrule
  \end{tabular}
    \vspace{-5pt}
  \caption{\textbf{Quantitative comparisons with text-to-3D content and text-to-3D human methods.} }
  \label{tab:fid_clip_score}
  \vspace{-10pt}
\end{table}


\subsection{Quantitative Evaluation}
Evaluating the quality of generated 3D models quantitatively can be challenging. However, we attempt to assess HumanNorm using two specific metrics. Firstly, we compute the Fréchet Inception Distance (FID)~\cite{heusel2017gans}, a measure that compares the distribution of two image datasets. In our case, we calculate the FID between the views rendered from the generated 3D humans and the images produced by Stable Diffusion V1.5~\cite{rombach2022high}. In total, 30 prompts are used and 120 images are rendered or generated for each prompt. Secondly, we utilize the CLIP score~\cite{hessel2021clipscore} to measure the compatibility between the prompts with the rendered views of 3D humans. The results are detailed in \cref{tab:fid_clip_score}. As can be observed, HumanNorm achieves a lower FID score. This suggests that the views rendered from our 3D humans are more closely aligned with the high-quality 2D images generated by the stable diffusion model. Furthermore, the superior CLIP score of HumanNorm indicates our enhanced capability to generate humans that are more accurately aligned with the prompts. Finally, we also conduct a user study to evaluate HumanNorm. The details of this study are provided in our supplementary.

\begin{figure}[t]
\includegraphics[width=\linewidth]{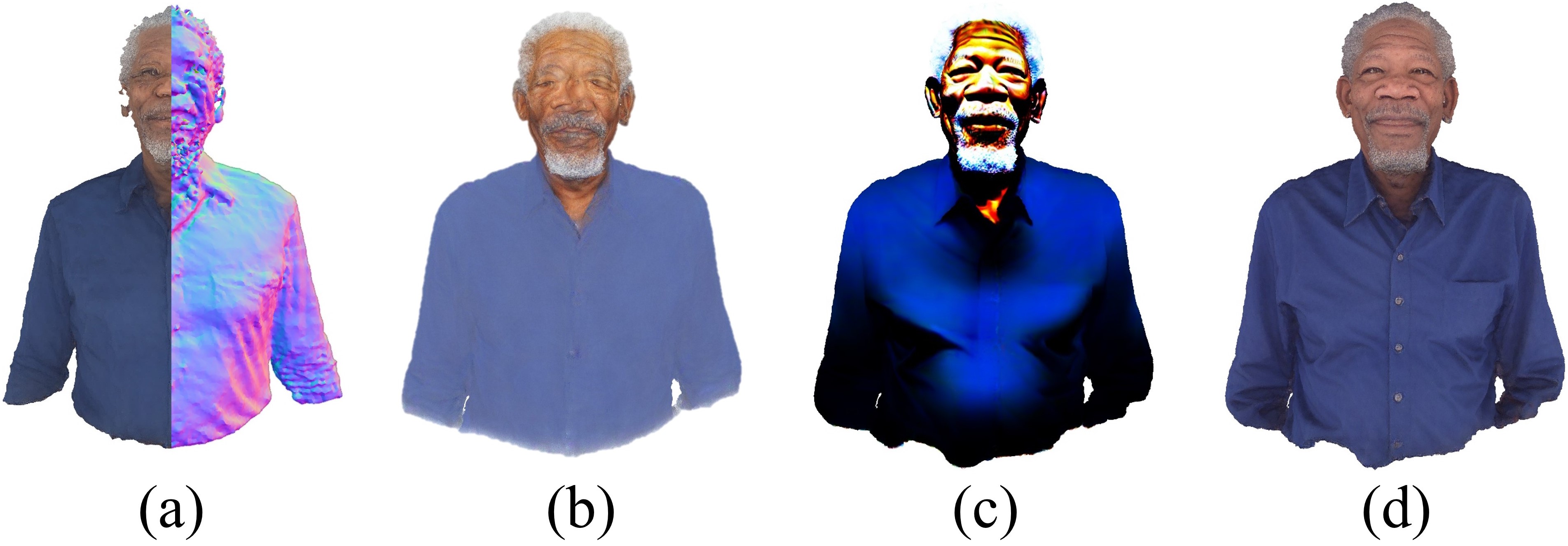}
\caption{\textbf{Ablation studies.} (a) Without normal-adapted and depth-adapted diffusion. (b) Without normal-aligned diffusion model. (c) Without multi-step SDS loss. (d) The full method.}
\label{fig:ablations1}
\end{figure}

\begin{figure}[th]
\includegraphics[width=\linewidth]{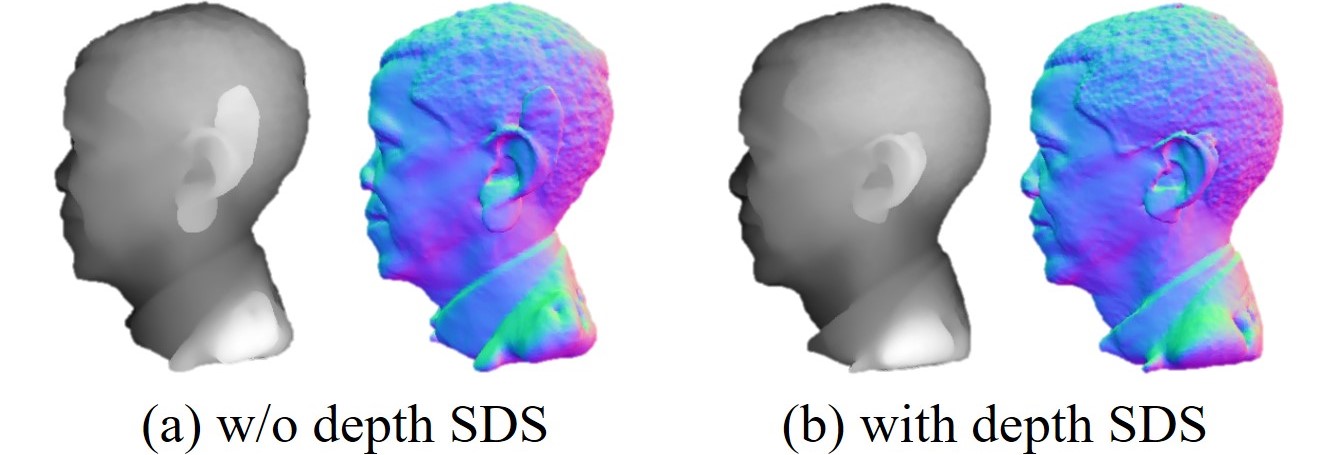}
\caption{\textbf{Importance of depth SDS.}}
\label{fig:ablation_text2depth}
\end{figure}

\subsection{Ablation Studies}

\noindent\textbf{Effectiveness of normal-adapted and depth-adapted diffusion models.} In ~\cref{fig:ablations1}~(a), we show the geometry generated by a text-to-image diffusion model instead of our normal-adapted and depth-adapted diffusion models. One can see that the method struggles to generate facial geometry, and holes appear on ears. Additionally, the results display smoother clothing wrinkles. The experiment demonstrates that our normal-adapted and depth-adapted diffusion models are beneficial in generating high-quality geometry. 

\noindent\textbf{Effectiveness of normal-aligned diffusion model.} In ~\cref{fig:ablations1}~(b), we experiment with the removal of the normal-aligned diffusion model, opting instead for a text-to-image diffusion model for texture generation. The resulting texture, as can be observed, is somewhat blurry and fails to accurately display geometric details. This is because the text-to-image diffusion model struggle to align the generated texture with geometry. However, using the normal-aligned diffusion model, our method manages to overcome these limitations. It achieves more precise and intricate details, leading to a significant enhancement for the appearance of the 3D humans.

\noindent\textbf{Effectiveness of multi-step SDS loss.} In ~\cref{fig:ablations1}~(c), we present the result generated when only the SDS loss is used in the texture generation. The generated model is noticeably over-saturated. However, as shown in ~\cref{fig:ablations1}~(d), the texture generated through multi-step SDS loss exhibits a more realistic and natural color, which underscores the effectiveness of the multi-step SDS loss. 

\noindent\textbf{Effectiveness of depth SDS.} Since normal maps lack depth information, optimizing geometry by only calculating normal SDS loss may lead to failed geometry in some regions. As shown in ~\cref{fig:ablation_text2depth}~(a), the ear exhibits artifacts when only using normal SDS loss. This is because the normal of the artifacts is similar to the normal of the head, making it non-salient for the normal diffusion model. In contrast, we can clearly see the artifacts in the depth map. In ~\cref{fig:ablation_text2depth}~(b), it’s evident that the artifacts are reduced when adding the additional depth SDS loss based on our depth-adapted diffusion model, which demonstrates the effectiveness of introducing depth SDS.

\begin{figure}[!t]
\includegraphics[width=\linewidth]{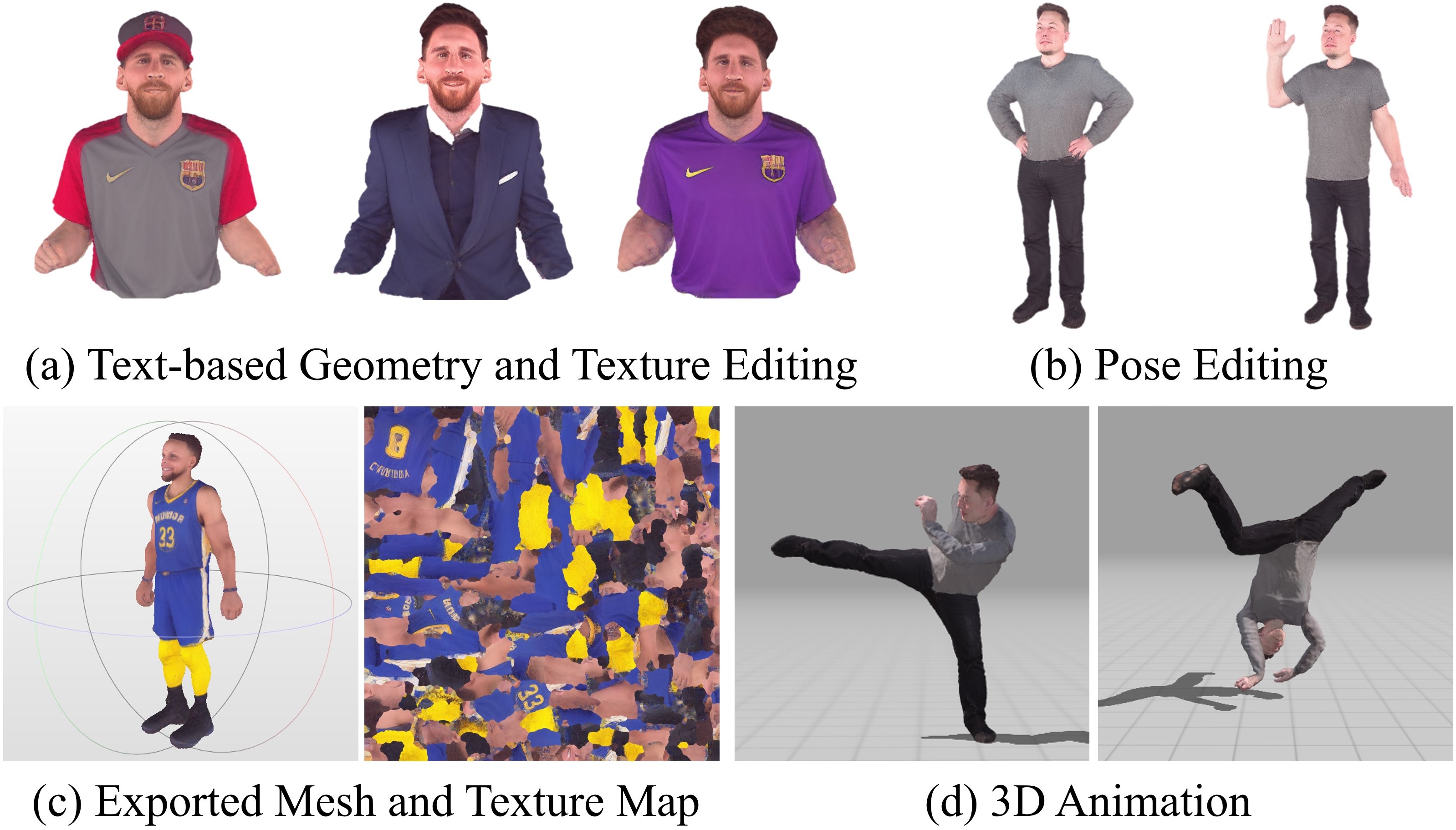}
\caption{\textbf{Applications of HumanNorm.}}
\label{fig:applications}
\end{figure}

\subsection{Applications}
\noindent\textbf{Text-based Editing.} HumanNorm offers the capability to edit both the texture and geometry of the generated 3D humans by adjusting the input prompt. As demonstrated in \cref{fig:applications}~(a), we modify the color and style of Messi’s clothing, as well as his hairstyle.

\noindent\textbf{Pose Editing.} HumanNorm also provides the ability to edit the pose of generated 3D humans by adjusting the pose of the mesh used for initialization and modifying the prompts. The results of pose editing are displayed in \cref{fig:applications}~(b).

\noindent\textbf{3D Animation.} HumanNorm enables the creation of lifelike human mesh featuring about 400K distinct faces and intricate 2K-resolution texture map. Based on the high-quality models, we can animate them using full-body motion sequences. Results are presented in \cref{fig:applications}~(c-d)




\section{Conclusion}
We presented HumanNorm, a novel method for high-quality and realistic 3D human generation. By learning the normal diffusion model, we improved the capabilities of 2D diffusion models for 3D human generation. Utilizing the trained normal diffusion model, we introduced a diffusion-guided 3D generation framework. Additionally, we devised the progressive strategy for robust geometry generation and the multi-step SDS loss to address the over-saturation problem. We demonstrated that HumanNorm can generate 3D humans with intricate geometric details and realistic appearances, outperforming existing methods.

\noindent\textbf{Limitations and future work.} HumanNorm primarily focuses on addressing the geometric and textural challenges present in existing methods. As a result, 3D humans generated by HumanNorm necessitate a rigged human skeleton for 3D animation. In our future work, we plan to incorporate SMPL-X to directly animate 3D humans and improve the quality of body details such as fingers. Additionally, our generated texture may exhibit undesired shading. To address this, we are considering the use of Physically-Based Rendering (PBR) for material estimation and relighting. 

{
    \small
    \bibliographystyle{ieeenat_fullname}
    \bibliography{main}
}
\appendix
\clearpage
\setcounter{page}{1}
\maketitlesupplementary

\section{3D representations}
\noindent\textbf{SDF representation.} Signed Distance Fields (SDF) is a 3D representation used to describe the geometry surface of an object. It is expressed implicitly through neural networks like MLP. For a sampling point x, everything satisfying $f(x) = 0$ is considered to be part of the object's surface, while the region where $f(x) < 0$ represents the object's interior, and $f(x) > 0$ indicates the object's exterior. SDF can be employed in the synthesis of images from arbitrary viewpoints through methods such as differentiable volume rendering or differentiable marching cubes for geometry extraction and re-rendering. 


\noindent \textbf{DMTET representation.} DMTET~\cite{shen2021deep} is a hybrid 3D representation that combines explicit and implicit forms. It divides 3D space into dense tetrahedra, which is an explicit partition. Simultaneously, the vertices of these tetrahedra record properties of the 3D object, including SDF, deformation, color, etc. These properties are expressed through the implicit functions of neural networks. By combining explicit and implicit representations, DMTET can be optimized more efficiently and easily transformed into explicit structures like mesh representations. During the generation process, DMTET can be converted into a mesh in a differentiable manner, enabling rapid high-resolution multi-view rendering. We utilize DMTET as the 3D representation in both the geometry generation and texture generation phases.

\section{Implementation Details}

\textbf{Dataset.} Our dataset comprises 2952 3D human body models. These include 526 models from the THuman2.0 dataset~\cite{yu2021function4d}, 1779 models from the Twindom dataset~\cite{TwinDom}, and 647 models from the CustomHumans dataset~\cite{ho2023learning}. We use these models to generate depth maps, normal maps, and color maps. To augment the dataset, we divide the human body into four distinct sections: the head, the upper body, the lower body, and the full body. For each model, we render a set of 120 images, each set comprising depth maps, normal maps, and color maps. The normal maps are transformed into camera coordinates by the rotation of the camera parameter. We utilize CLIP~\cite{radford2021learning} to generate prompts for the images, supplementing them with additional text to label various data types such as ``depth map" and ``normal map". We also include view-dependent descriptors for the view direction, such as ``front view", ``back view", ``left side view", and ``right side view", as well as body-aware text for specific regions of the human body, including ``head only", ``upper body", ``lower body", and ``full body".

\noindent\textbf{Training of normal-adapted and depth-adapted diffusion models.} The base stable diffusion model used in our method is Stable Diffusion V1.5~\cite{rombach2022high}. We fine-tune the stable diffusion model using our depth pairs and normal pairs for 15K iterations. The learning rate is set to $1\times10^{-5}$ and the batch size is set to 4. Exponential Moving Average (EMA) is used during the training.  After fine-tuning, we obtain a normal-adapted diffusion model and a depth-adapted diffusion model. The fine-tuning code is from Diffusers (\url{https://huggingface.co/docs/diffusers/index}), a library for state-of-the-art pretrained diffusion models for generating images, audio, and even 3D structures of molecules

\noindent\textbf{Training of normal-aligned diffusion model.} To guide the generation of stable diffusion using a normal map, we follow the fine-tuning strategy of ControlNet~\cite{zhang2023adding}. We fine-tune Stable Diffusion V1.5 for 30K iterations using normal-image pairs. The normal maps are used as extra conditions. The learning rate is set to $1\times10^{-5}$ and the batch size is set to 4. The fine-tuning code is also from Diffusers.

\noindent\textbf{Details of progressive positional encoding.} In progressive geometry generation, we employ progressive positional encoding. Specifically, the position encoding for SDF features in DMTET has a total of 32 dimensions, where the lower dimensions represent lower-frequency features and higher dimensions represent higher-frequency features. Initially, we utilize a 32-dimensional mask with the first 16 dimensions set to 1 and the latter 16 dimensions set to 0. We multiply this mask with the SDF's position encoding to remove the high-frequency components. During training, every 500 iterations, we convert 2 of the 0 positions in the mask to 1, gradually enabling the network to learn high-frequency components. After 4,000 iterations, all positions in the mask become 1, resulting in the position encoding encompassing both low-frequency and high-frequency components.

\begin{figure*}[!t]
\includegraphics[width=\textwidth]{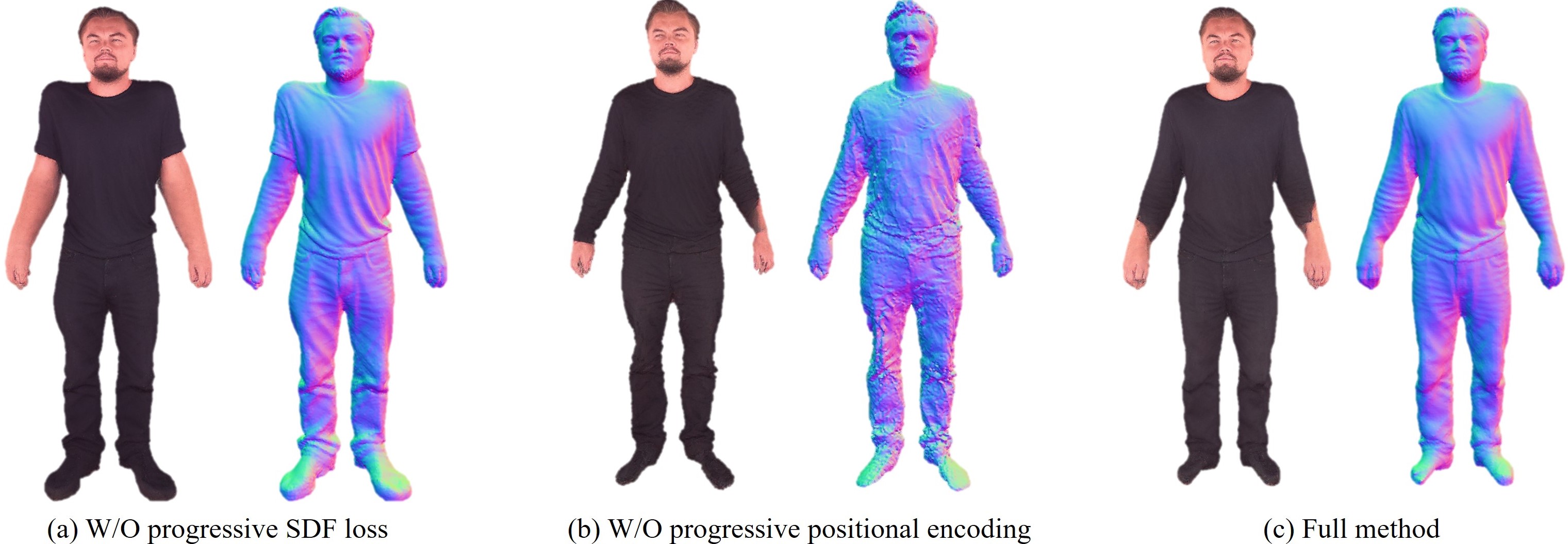}
\caption{\textbf{Importance of progressive SDF loss and progressive positional encoding.} }
\label{fig:ablation_supp}
\end{figure*}

\noindent\textbf{Details of progressive SDF loss.} During the training process, at the 3,000 iterations, we extract the current geometry to form a coarse mesh. This coarse mesh exhibits a reasonable shape and features a relatively smooth surface. We utilize it to compute the SDF loss for subsequent stages. Specifically, within the bounding box of the 3D generation, we randomly sample 100,000 points at each iteration. Then we calculate the SDF loss by comparing the SDF values of these points in the coarse mesh with the SDF values predicted by the network. The weight of the SDF loss among all the losses is set to 1500 and is only computed after the 3,000 iterations.

\noindent\textbf{Details of geometric resolution.} 
We similarly adopt an approach to gradually increase the geometric resolution. Initially, the resolution of the DMTET in 3D space is set to $128^3$. As training proceeds, we incrementally double this resolution every 3,000 iterations. So at 3,000 iterations, the resolution is set to $256^3$, and it will eventually reach $512^3$ at 6,000 iterations. In the early training stages, this results in fewer generated geometry facets, with each facet occupying more pixels in the rendered images. Consequently, the gradients produced by the loss are more evenly distributed across the points of each facet, leading to more stable geometry generation. As the geometric resolution increases, the number of geometry facets also increases, allowing for the representation of more intricate details, including features like hair and clothing folds.

\noindent\textbf{Details of texture generation.} In texture generation, the initial 2,000 iterations are utilized as coarse-level optimization and employ SDS loss, while the subsequent 8,000 iterations serve as fine-level optimization, using the multi-step SDS loss and perceptual loss. For the multi-step SDS loss, the diffusion model performs varying numbers of iterations based on the timestep $t$ with added noise. Specifically, The total timestep of our diffusion model is 1000, when the timestep is $t$, the diffusion model iterates $(t / 25 + 1)$ times. We employ the DPM++ solver~\cite{lu2022dpm} as our diffusion scheduler. To enhance training stability, we also incorporate a DU (Dataset Update) strategy similar to what was proposed in Instruct-NeRF2NeRF~\cite{haque2023instruct}. During computation for the multi-step loss at each iteration, we save the image results of multi-step diffusion denoising in a cached dataset, which are reused in subsequent training processes. Every 10 iterations, we will use multi-step diffusion denoising to update the images in the cached dataset.

\noindent\textbf{Noise and guidance scale of the diffusion model.} In the geometry stage, both our normal-adapted and depth-adapted diffusion models have a guidance scale of 50. Similar to the strategy employed in progressive geometry generation, we introduce noise progressively during the geometry stage. In the first 5,000 steps, the timestep $t$ of noise follows the distribution $\mathcal{U}(0.02, 0.8)$. Between 5,000 and 8,000 steps, the timestep $t$ of noise follows the distribution $\mathcal{U}(0.02, k)$ with parameter $k=0.2+(0.8-0.2)\frac{8000-step}{8000-5000}$. After 8,000 steps, the timestep $t$ of noise follows the distribution $\mathcal{U}(0.02, 0.2)$. In the texture stage, our geometry-guided diffusion model has a guidance scale of 7.5, and the controlled condition scale is set to 1.0. During the coarse level of texture generation, the timestep $t$ of noise follows the distribution $\mathcal{U}(0.02, 0.98)$. In the fine level, the timestep $t$ of noise follows the distribution $\mathcal{U}(0.02, 0.5)$.

\noindent\textbf{Learning rate and the weight of losses in 3D generation.} We adopt the AdamW optimizer in 3D generation. The learning rate of $\theta_g$ is set to $2\times10^{-5}$ and the learning rate of $\theta_c$ is set to $1\times10^{-3}$. In the geometry generation, the weight of the normal SDS loss is set to $1.0$, and the weight of the depth SDS loss is $1.0$. In the texture generation, the weight of perceptual loss is set to $1.0$.


\begin{table*}[!t]
\caption{\textbf{Results of user study.} The table reports the user preference percentages in detail.}
\label{tab:user_study}
\begin{center}
\begin{tabular}{lcccccc}
\toprule
&\multicolumn{2}{c}{Q1 (\%)}  &\multicolumn{2}{c}{Q2 (\%)}  &\multicolumn{2}{c}{Q3 (\%)} \\
 \cmidrule(lr){2-3}  \cmidrule(lr){4-5} \cmidrule(lr){6-7} 
& Best  & Second best & Best  & Second best  & Most  & Second most  \\
\midrule
DreamFusion & 5.36  & 22.27 & 4.73  & 20.55 & 9.27  & 22.55 \\
LatentNeRF  & 3.09  & 11.82 & 6.64  & 8.45  & 8.45  & 12.91 \\
TEXTure     & 3.64  & 10.27 & 3.91  & 6.64  & 4.91  & 9.09  \\
Fantasia3D  & 9.91  & \textbf{41.45} & 10.45 & \textbf{50.55} & 12.64 & \textbf{39.00} \\
Ours        & \textbf{78.00} & 14.18 & \textbf{74.27} & 13.82 & \textbf{64.73} & 16.45 \\
\bottomrule \\
\end{tabular}

\begin{tabular}{lccc}
\toprule
&\multicolumn{1}{c}{Q1 (\%)}  &\multicolumn{1}{c}{Q2 (\%)}  &\multicolumn{1}{c}{Q3 (\%)} \\
\midrule
DreamHuman & 8.79 & 18.20 & 25.80 \\
TADA &16.91  & 11.25  & 15.20 \\
Ours       & \textbf{74.30} & \textbf{70.55} & \textbf{59.00} \\
\bottomrule
\end{tabular}

\end{center}
\end{table*}

\noindent\textbf{Part-based optimization.} We primarily divide the human body into four parts for generation: head, upper body, lower body, and the full body. To ensure that the rendered images cover each of these four parts separately, we predefine the camera positions and focal lengths accordingly. During the generation process, the probability of sampling from these four camera positions varies based on the optimization objective. When generating only the head, we sample from the camera capturing the head alone. When generating the upper body of the human, we assign a sampling probability of 0.7 to the upper body and 0.3 to the head. When generating the entire human body, we adjust the sampling strategy progressively. In the first 10,000 iterations, we assign a sampling probability of 0.7 to the entire body and 0.1 to each of the head, upper body, and lower body. In the subsequent 5,000 iterations, we assign a sampling probability of 0.1 to the entire body and 0.3 to each of the head, upper body, and lower body.

\section{User Study}
Following TADA~\cite{liao2023tada} and DreamHuman~\cite{kolotouros2023dreamhuman}, we conducted a user study to further assess the quality of the 3D human models generated by our method. Our approach was compared with five state-of-the-art methods across 30 prompts. For each prompt, 50 volunteers (comprising 40 students specializing in computer vision and graphics, and 10 members of the general public) evaluated the color and normal map videos rendered from the generated 3D humans. They voted on three questions:
\begin{itemize}
\item Q1: Which 3D human model exhibits the best (and second best) texture quality? 
\item Q2: Which 3D human model displays the best (and second best) geometric quality? 
\item Q3: Which 3D human model aligns most closely (and second most closely) with the given prompt? 
\end{itemize}
Since the source code of DreamHuman~\cite{kolotouros2023dreamhuman} is not publicly accessible, we sourced the results from its project page. The results of LatentNeRF~\cite{metzer2023latent}, TEXTure~\cite{richardson2023texture}, Fantasia3D~\cite{chen2023fantasia3d}, and TADA~\cite{liao2023tada} are produced using their official code with default settings. Meanwhile, the results of DreamFusion~\cite{poole2023dreamfusion} are generated using an unofficial implementation in ThreeStudio, a unified framework for 3D content creation (\url{https://github.com/threestudio-project/threestudio}). We all collect 1,500 pairwise comparisons. The results are shown in ~\cref{tab:user_study}. One can see that our method surpasses the performance of the text-to-3D content methods and text-to-3D human methods, particularly in terms of geometry and texture quality. These results underscore the superior performance of our approach.

\section{More Comparisons}
\begin{figure*}[th]
\includegraphics[width=\textwidth]{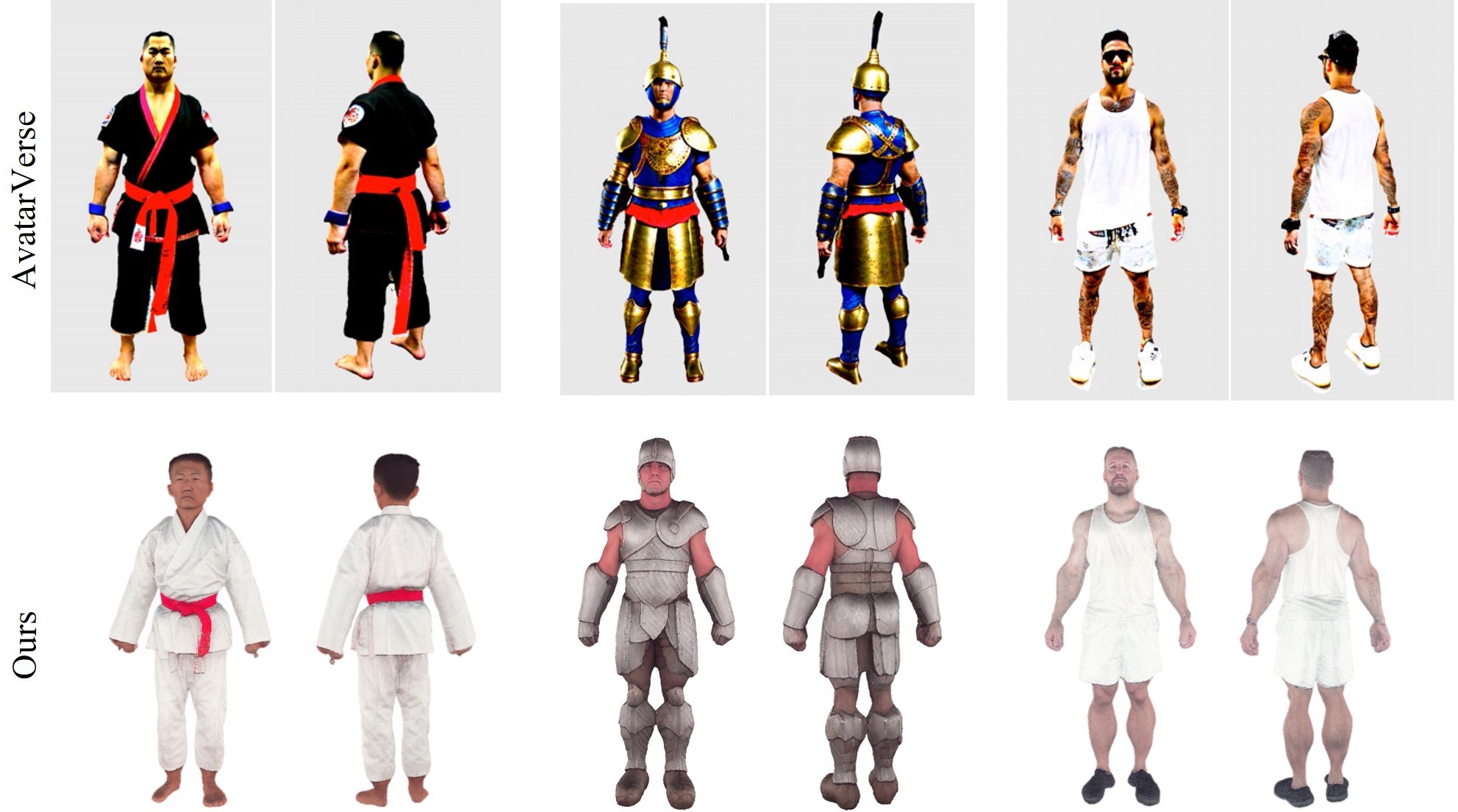}
\caption{\textbf{Comparisons with AvatarVerse.} The results of AvatarVerse are copied from its paper. }
\label{fig:avatarverse}
\end{figure*}

\begin{figure*}[th]
\includegraphics[width=\textwidth]{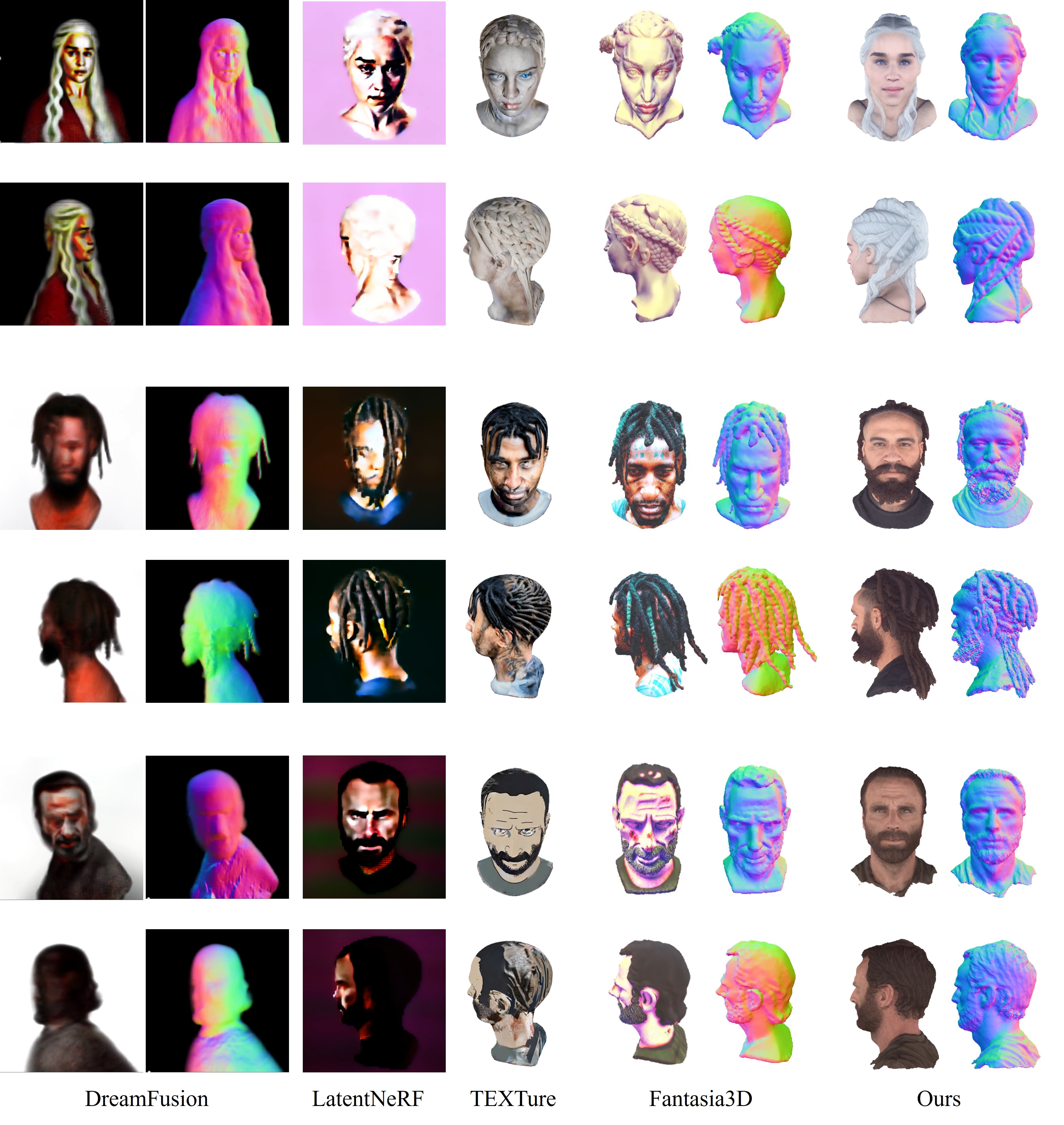}
\caption{\textbf{Comparison with text-to-3D content methods on the head-only 3D human generation.} }
\label{fig:supp_head}
\end{figure*}

\begin{figure*}[th]
\includegraphics[width=\textwidth]{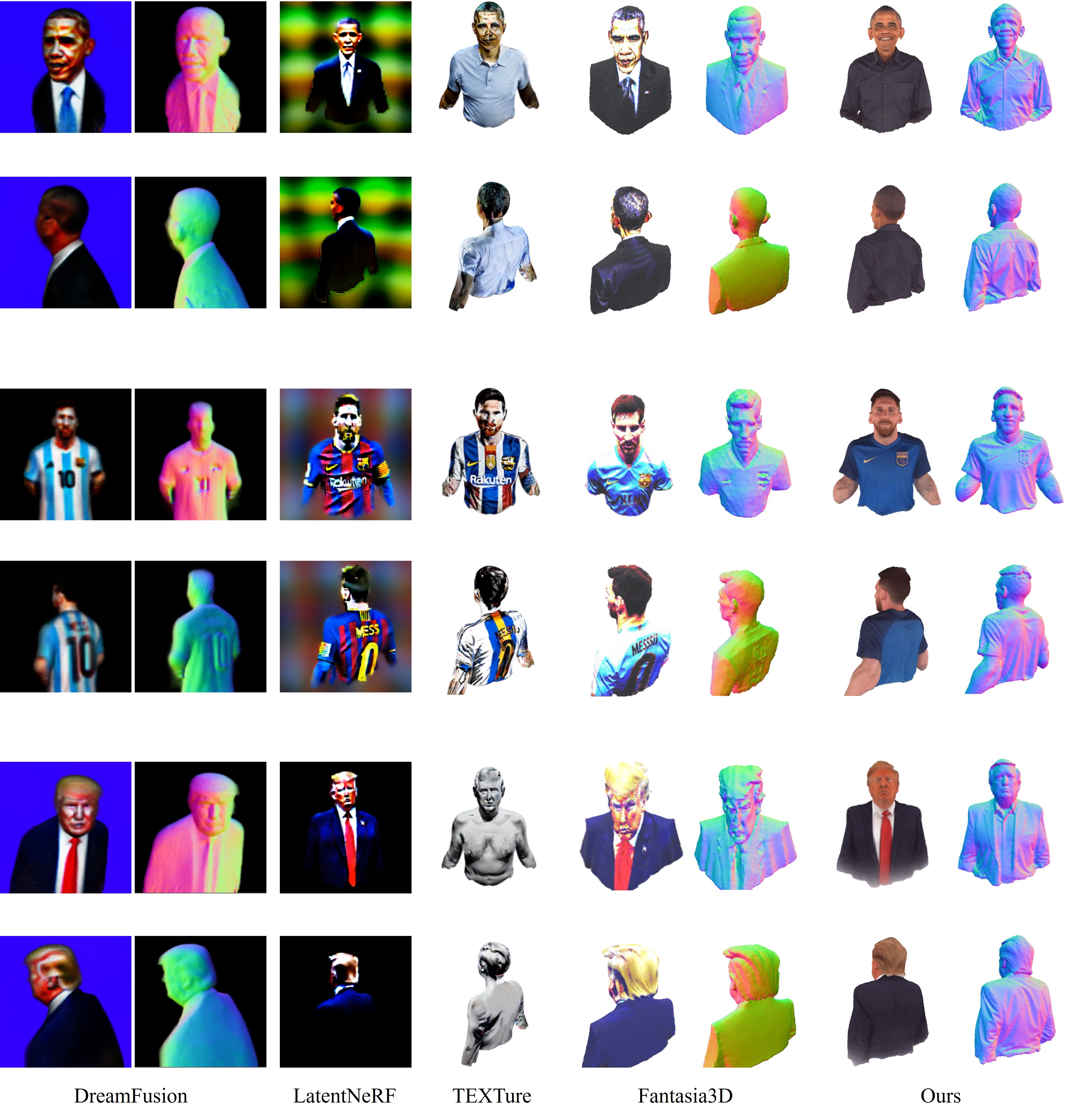}
\caption{\textbf{Comparison with text-to-3D content methods on the upper-body 3D human generation.}  }
\label{fig:supp_upper}
\end{figure*}

\begin{figure*}[th]
\includegraphics[width=\textwidth]{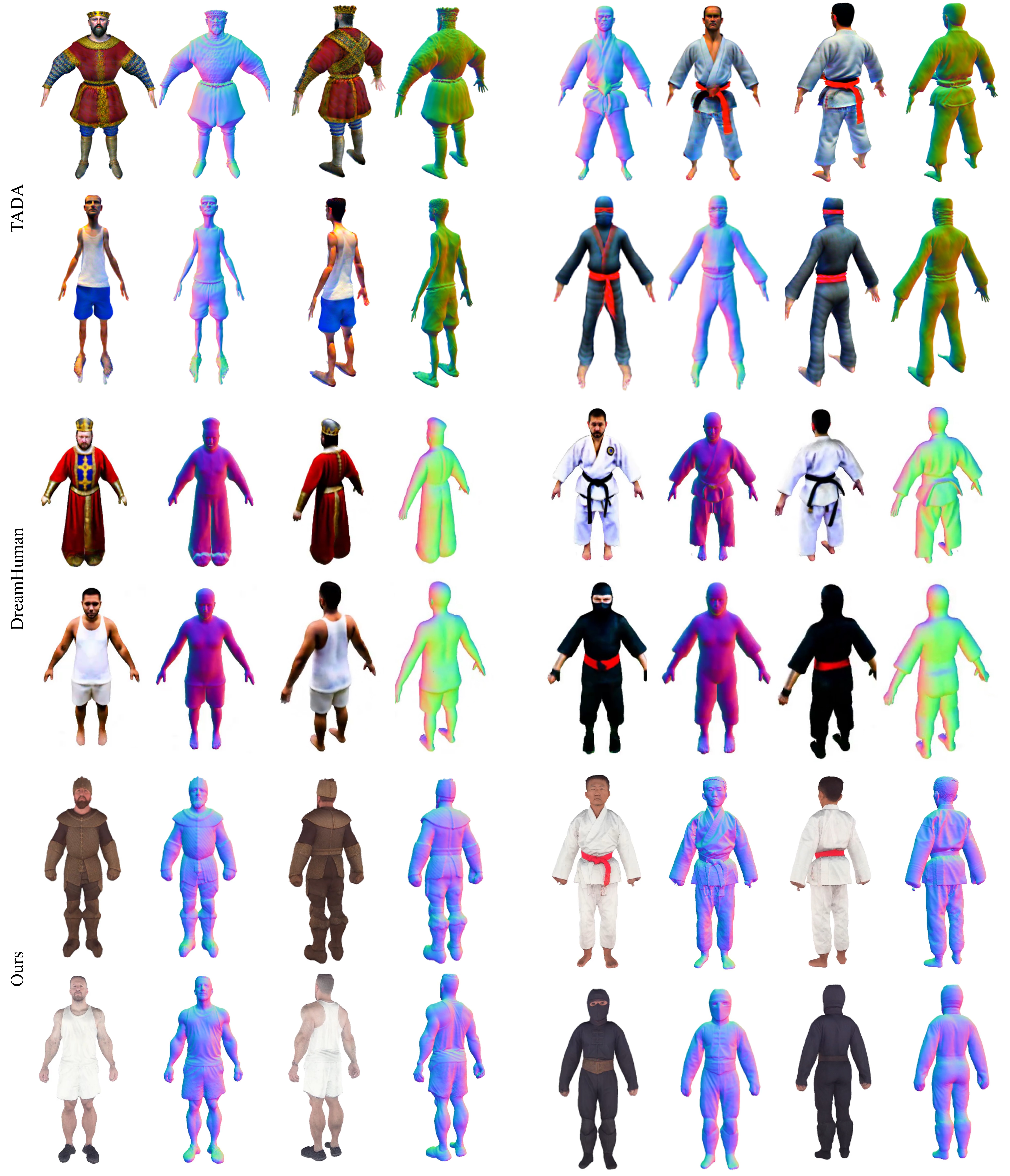}
\caption{\textbf{Comparison with text-to-3D human methods on the full-body 3D human generation. }}
\label{fig:supp_full}
\end{figure*}

We offer further qualitative comparisons with the four text-to-3D content methods and the two text-to-3D human methods. As depicted in \cref{fig:supp_head} and \cref{fig:supp_upper}, Fantasia3D may generate textures that are not aligned with the geometry (as seen in the second row of \cref{fig:supp_head}). However, the textures produced by our method are accurately aligned with the generated geometry. When compared to the four text-to-3D content methods, our method can generate head-only and upper-body 3D humans with more detailed geometry and a more realistic appearance. In \cref{fig:supp_full}, we present full-body results in comparison with DreamHuman and TADA. It is evident that the results produced by baselines contain over-saturated textures and smooth geometry, whereas our method yields a more natural appearance and geometric details. Additionally, we add a comparison with AvatarVerse~\cite{zhang2023avatarverse}, as shown in \cref{fig:avatarverse}. The 3D humans by AvatarVerse are over-saturated. In contrast, HumanNorm produces results with appearances that are more lifelike. 



\section{More Ablation Studies}

\textbf{Effectiveness of progressive SDF loss.} In \cref{fig:ablation_supp}~(a), we display the results obtained in the absence of progressive SDF loss. The 3D human exhibits a distorted body shape. However, the introduction of progressive SDF loss effectively constrains the wrong growth of the human body, thereby avoiding unreasonable body shapes.

\noindent\textbf{Effectiveness of progressive positional encoding.} In \cref{fig:ablation_supp}~(b), we conduct an experiment where the frequency of hash encoding is fixed. The results reveal extensive noise on the surface of the geometry, which can be attributed to the high-frequency content learned during the initial training phase. A contrasting case is presented in \cref{fig:ablation_supp} (c) when a progressive positional encoding is employed. Our strategy reduces the learning of high-frequency information during the initial training phase, resulting in a stable geometry devoid of geometric noise.

\section{Applications}
\begin{figure*}[th]
\includegraphics[width=\textwidth]{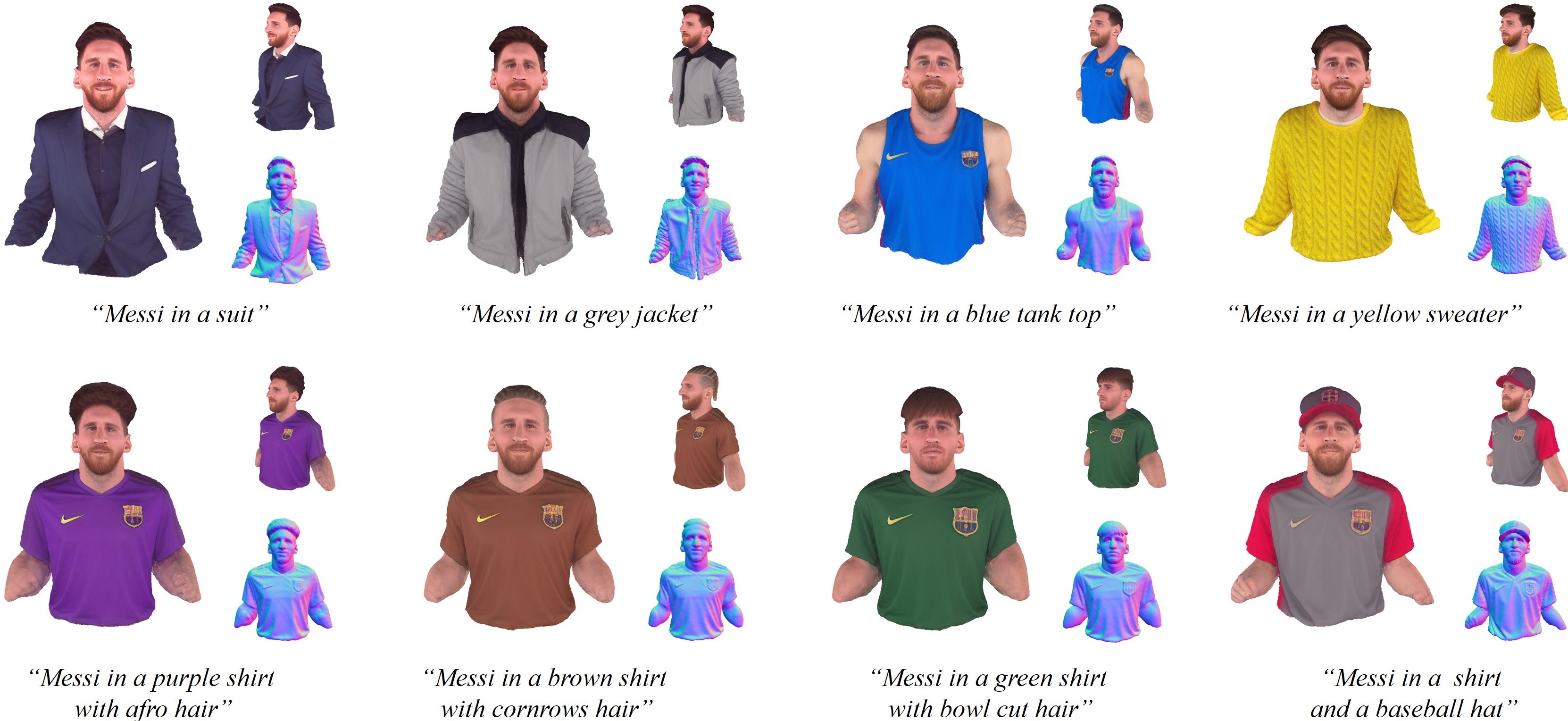}
\caption{\textbf{Text-based editing.} Our method provides the ability to modify both the texture and geometry of the generated 3D humans by simply altering the input prompt. }
\label{fig:editing}
\end{figure*}

Our method offers the capability to edit both the texture and geometry of the generated 3D humans by adjusting the input prompt. As demonstrated in \cref{fig:editing}, we modify the color and style of Messi’s clothing, as well as his hairstyle, all while maintaining his identity. While geometry editing poses a greater challenge than texture editing, our method exhibits precise control over geometry generation, even allowing us to generate Messi wearing a hat. Furthermore, the edited geometry is rich in detail, as evidenced by the intricate details in the sweater. \textbf{More applications can be viewed on our attached project page.}



\section{Ethics statement}
The objective of HumanNorm is to equip users with a powerful tool for creating realistic 3D Human models. Our method allows users to generate 3D Humans based on their specific prompts. However, there is a potential risk that these generated models could be misused to deceive viewers. This problem is not unique to our approach but is prevalent in other generative model methodologies. Moreover, it is of paramount importance to give precedence to diversity in terms of gender, race, and culture. As such, it is essential for current and future research in the field of generative modeling to consistently address and reassess these considerations.

\end{document}